\documentclass[conference]{IEEEtran}
\IEEEoverridecommandlockouts
\usepackage{amsmath,amssymb,amsfonts}
\usepackage[compress]{cite}
\usepackage{balance}
\usepackage{algorithmic}
\usepackage{algorithm}
\usepackage{graphicx}
\usepackage{subcaption}
\usepackage{textcomp}
\usepackage{pgfplots}
\usepackage{tikz}

\usepackage[utf8]{inputenc} 
\usepackage[T1]{fontenc} 
\def\BibTeX{{\rm B\kern-.05em{\sc i\kern-.025em b}\kern-.08em
    T\kern-.1667em\lower.7ex\hbox{E}\kern-.125emX}}

\newcommand{\R}{\mathbb{R}}

\definecolor{green_graph}{RGB}{85, 168, 104}
\definecolor{blue_graph}{RGB}{76, 114, 176}
\definecolor{gray_graph}{RGB}{136, 138, 133}

\begin{document}

\title{Feature-based Event Stereo Visual Odometry
}

\author{Antea Hadviger$^{1}$, Igor Cvi\v{s}i\'{c}$^{1}$, Ivan Markovi\'{c}$^1$, Sacha Vra\v{z}i\'{c}$^{2}$, Ivan Petrovi\'{c}$^1$
\thanks{
$^{1}$ Antea Hadviger, Igor Cvi\v{s}i\'{c}, Ivan Markovi\'{c}, and Ivan Petrovi\'{c} are with the University of Zagreb
Faculty of Electrical Engineering and Computing, Laboratory for Autonomous Systems and Mobile Robotics, Croatia. {\{antea.hadviger, igor.cvisic, ivan.markovic, ivan.petrovic\}@fer.hr}
}
\thanks{
$^{2}$ Sacha Vra\v{z}i\'{c} is with Rimac Automobili, Croatia.  {sacha.vrazic@rimac-automobili.com}
}
}

\maketitle

\begin{abstract}
Event-based cameras are biologically inspired sensors that output \textit{events}, i.e., asynchronous pixel-wise brightness changes in the scene.
Their high dynamic range and temporal resolution of a microsecond makes them  more reliable than standard cameras in environments of challenging illumination and in high-speed scenarios, thus developing odometry algorithms based solely on event cameras offers exciting new possibilities for autonomous systems and robots.
In this paper, we propose a novel stereo visual odometry method for event cameras based on feature detection and matching with careful feature management, while pose estimation is done by reprojection error minimization.
We evaluate the performance of the proposed method on two publicly available datasets: MVSEC sequences captured by an indoor flying drone and DSEC outdoor driving sequences.
MVSEC offers accurate ground truth from motion capture, while for DSEC, which does not offer ground truth, in order to obtain a reference trajectory on the standard camera frames we used our SOFT visual odometry, one of the highest ranking algorithms on the KITTI scoreboards.
We compared our method to the ESVO method, which is the first and still the only stereo event odometry method, showing on par performance on the MVSEC sequences, while on the DSEC dataset ESVO, unlike our method, was unable to handle outdoor driving scenario with default parameters.
Furthermore, two important advantages of our method over ESVO are that it adapts tracking frequency to the asynchronous event rate and does not require initialization.
\end{abstract}

\begin{IEEEkeywords}
event-based cameras, dynamic vision sensors, stereo vision, visual odometry
\end{IEEEkeywords}


\section{Introduction} \label{sec:intro}

\begin{figure}[!t]
	\centering
	\includegraphics[width=0.75\linewidth]{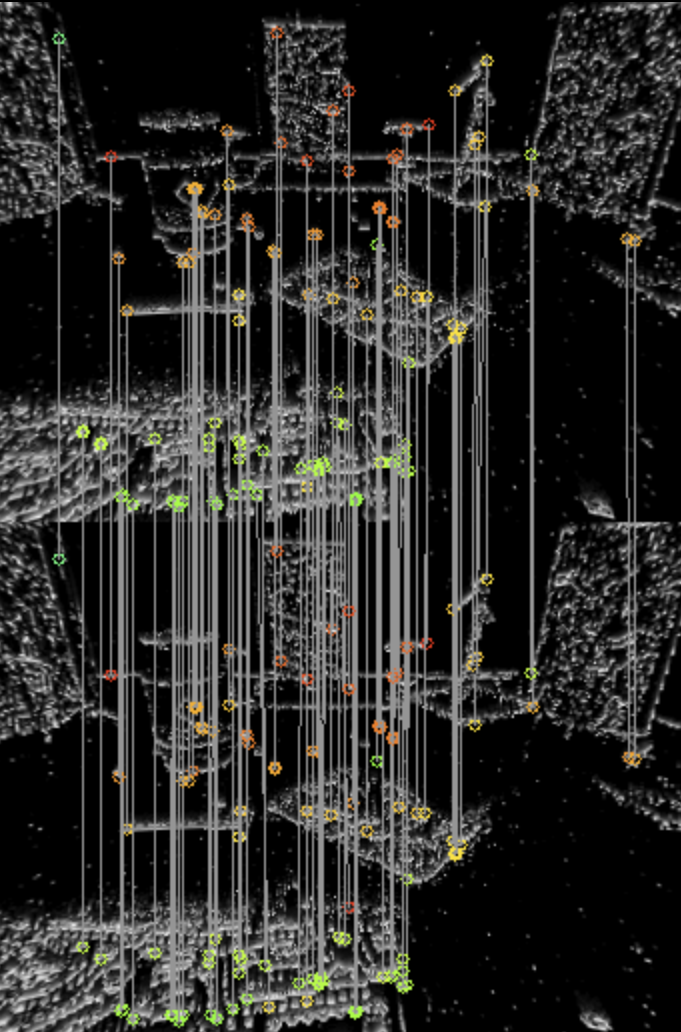}
	\caption{Event time surface $\tau$ samples from a MVSEC indoor sequence with a subset of temporally matched features between $\tau_R(t)$ and $\tau_R(t-\delta t)$ used for pose optimization (color indicates depth).}
	\label{fig:ts}
\end{figure}

Event cameras, also referred to as dynamic vision sensors (DVS), are biologically inspired sensors that detect pixel-wise changes in brightness intensity and asynchronously report them as \textit{events} at the time of their occurence, as opposed to capturing images of the whole scene with a fixed frame-rate.
Many advantages over standard frame-based cameras, like high temporal resolution of a microsecond, low latency, high dynamic range of 120~dB, and low power consumption, make them a very interesting sensor in challenging scenarios, including scenes with dynamic illumination or high speed motion, since they do not suffer from motion blur like standard cameras.
Given that the output of event cameras is fundamentally different than the standard camera frames, unleashing the full potential of the sensor depends on developing novel asynchronous processing methods, but they have already been proven as a valuable asset for various robotic perception tasks, from depth estimation to simultaneous localization and mapping (SLAM) \cite{gallego2019event}.

Event-based localization solutions have been offered for problems of gradually increasing complexity, with restrictions in terms of dimensionality, type of scene, and type of motion.
First attempts tackled strictly rotational \cite{kim2008simultaneous} or planar motions \cite{weikersdorfer2013simultaneous}.
An adaptation of the semi-direct visual odometry algorithm \cite{forster2014svo} for event cameras was proposed in \cite{kueng2016low} to track 6-DoF motion in natural scenes.
However, this method is not purely event-based since it detects features and grayscale frames, and tracks them asynchronously using the event stream.
A method for 6-DOF tracking in a known environment described by a photometric 3D map of the scene was presented in \cite{bryner2019event}. This approach estimates the camera pose by comparing raw events to the predictions of events that would be generated according to the known map of the scene and the candidate camera motion.
A purely event-based 6-DoF visual odometry for natural scenes was presented in \cite{kim2016real}. The method uses three interleaved probabilistic filters to process the information from a single event camera, performing pose tracking, depth estimation and intensity estimation. However, the method is computationally expensive, and thus requires a GPU to achieve real-time performance.
The system proposed in \cite{rebecq2016evo} tackles the same SLAM problem, but proposes to solve it without recovering image intensity to estimate depth, thus reducing computational complexity, while also eliminating the propagation of intensity estimation error.
This method was based on a purely geometric approach to recover semi-dense 3D structure of the scene that leverages the fact that event cameras naturally respond to edges \cite{rebecq2016emvs}.

Since many event cameras, e.g. DAVIS, have the inertial measurement unit (IMU) readily integrated, several visual-inertial odometry approaches have been proposed.
A method that tracks features in the event stream using an expectation-maximization scheme that estimates the optical flow was proposed in \cite{zhu2017event1}. Pose estimates are then produced by an extended Kalman filter that fuses the feature tracks and the inertial measurements.
A similar approach was presented in \cite{rebecq2017real}, which generates motion-compensated event frames using the optical flow generated by the camera's motion to extract the features and track them with a pyramidal Lucas-Kanade tracker.
The tracks are combined with inertial information using keyframe-based nonlinear optimization.
The authors also extended the method to leverage the complementary advantages of both standard and event cameras in combination with an IMU by tracking features using standard frames and events independently, and feeding the feature tracks to the same optimization module \cite{vidal2018ultimate}.
The first, and currently the only other stereo event-based visual odometry method, referred to as ESVO, was proposed in \cite{zhou2021event}.
The system follows a parallel tracking-and-mapping approach, where both modules operate in an interleaved manner estimating both ego-motion and the semi-dense 3D structure of the scene.
The mapping module fuses depth estimates from multiple viewpoints obtained by leveraging spatio-temporal consistency of events, while the camera poses are estimated by solving an event registration problem.
The authors' implementation of this algorithm is open-sourced, which we make use of in order to compare it to our method.

In this paper, we propose a first event stereo visual odometry method based on features for 6-DoF motion and natural scenes, inspired by traditional frame-based algorithms, but adapted for asynchronous event data.
This allows for the use of event cameras in scenarios where frame-based odometries might be unreliable.
Our method estimates camera ego-motion by minimizing the reprojection error of a carefully selected set of features.
The features are stereo and temporally matched through consecutive left and right event time surfaces generated asynchronously at arbitrary timestamps, as seen in Fig.~\ref{fig:ts}.
Our approach does not rely on a predefined frequency, but adapts it to the number of incoming events and also does not require any initialization.
We evaluate the proposed method on two different publicly available datasets and compare it to ESVO where applicable.

The rest of the paper is organized as follows.
Section \ref{sec:method} presents our proposed stereo visual odometry method consisting of four main steps: event representation, feature detection, feature matching and pose optimization.
In Section \ref{sec:results} we present the experimental results and we conclude the paper in Section \ref{sec:conclusion}.

\section{Proposed Event Stereo Visual Odometry} \label{sec:method}

\begin{figure*}[!t]
	\centering
	\includegraphics[width=0.9\linewidth]{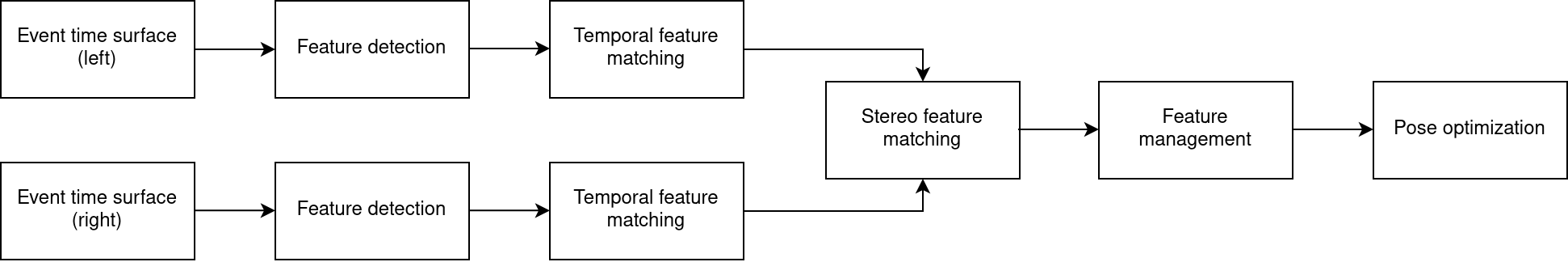}
	\caption{Pipeline of the proposed feature-based event stereo odometry framework that follows the same paradigm as traditional frame-based methods, but each module is adapted for asynchronous event data.}
	\label{fig:pipeline}
\end{figure*}

The proposed event stereo visual odometry method relies on features and follows a traditional frame-based odometry framework.
However, as opposed to frame-based methods that work with a fixed frequency, the proposed method operates on an asynchronous event stream, therefore allowing for arbitrary frequency of the pose estimates.
We propose to have a dynamic frequency adapted to the number of events, to ensure that enough features can be detected for reliable estimation.
However, we also define maximum time between pose estimates to avoid stalling in case of too few events occurring.
An important advantage of our method is that it does not require any initialization, as opposed to ESVO that relies on successfully performing semi-global matching on event time surfaces for point cloud initialization, which can be unreliable at times.
Due to having fixed predefined mapping and tracking frequencies, ESVO also needs to perform reinitialization when there are not enough incoming events, while our method adapts to this kind of situation by yielding pose estimates with a lower frequency.
The pipeline of the proposed event stereo visual odometry method is presented in Fig. \ref{fig:pipeline}.
It relies on feature detection, tracking and matching, and calculates optimal pose by minimizing the reprojection error of the matched features.
In the sequel, we present our realizations of the method modules.

\subsection{Event Representation} \label{sec:repr}

An event camera outputs an asynchronous stream of events. An event $e_k = (\mathbf{x}_k, t_k, p_k)$ at the pixel $\mathbf{x}_k = (x_k, y_k)^\top$ and time $t_k$ is triggered as soon as the increment in brightness, i.e., log photocurrent $L = log(I)$, reaches a defined contrast threshold $C$:
\begin{equation}
	\Delta L(\mathbf{x}_k, t_k) = L(\mathbf{x}_k, t_k) - L(\mathbf{x}_k, t_k-\Delta t_k) = p_kC,
\end{equation}
where $\Delta t_k$ is the time since the last event at the given pixel, and $p_k \in \{+1, -1\}$ is the polarity of the brightness change, i.e., increase or decrease in brightness.
Our method does not operate on an event-per-event basis, but requires data representation that contains context and history of events.
A widely used data structure for event representation is a simple 2D map containing the timestamp $t_{last}(x, y)$ of last event that occurred at each pixel, for each polarity.
However, events might not be detected and transmitted perfectly reliably in some cases.
A sudden drastic change in brightness triggers a lot of events in the same pixel in a very short amount of time, but not all events can be registered due to hardware limitations, thus compromising the local ordering and representation of events, as shown in \cite{alzugaray2018asynchronous} for corner detection.
Therein, the authors propose to use a more restrictive data structure that aims to eliminate effects of noise and improve corner detection by introducing a new variable, the reference timestamp $t_{ref}(x, y)$.
They propose to always retrieve $t_{ref}$ instead of $t_{last}$, and update $t_{ref}$ only if an event was not triggered in the corresponding pixel during a fixed interval $\kappa$, i.e., $t > t_{last} + \kappa$, or if the polarity of the incoming event differs from the latest one.
These timestamps are used directly for feature detection in the Arc* algorithm as discussed in \ref{sec:detection}.

To construct feature descriptors, we use the timestamp values to construct an alternative event map commonly referred to as the \textit{time surface} (TS).
TSs are generated independently for left and right camera at an arbitrary time $t$ by converting the event timestamps to values between 0 and 1 using the exponential decay kernel, effectively emphasizing recent events over the past ones.
Namely, the value of TS for a location $\mathbf{x}$ at the time $t \geq t_{last}(\mathbf{x})$ is defined as:
\begin{equation}
\tau(\mathbf{x}, t)=\exp\bigg(-\frac{t-t_{last}(\mathbf{x})}{\delta}\bigg),
\end{equation}
where $\delta$ presents a constant decay rate, usually set to several tenths of milliseconds.
This kind of event map is representative of the scene structure as it emphasizes the edges, thus containing relevant information for building discriminative feature descriptors.

\subsection{Feature Detection} \label{sec:detection}

We use the publicly available implementation of the Arc* algorithm for event-based corner detection \cite{alzugaray2018asynchronous}.
This detector asynchronously performs corner detection on an event-per-event basis in real-time by leveraging the 2D map of event timestamps $t_{ref}$.
Upon the arrival of a new event, the algorithm inspects a circular set of events around its location and looks for corners, i.e. arc-like structures in the event timestamp map.
The used detector operates asynchronously on all incoming events, but we do not use all detections for pose estimation.
Only those corners whose timestamp is within a short interval from the currently observed time surface at the time $t$ are considered, in order to only retain detections which the time surface emphasizes the most.
Feature descriptors for the remaining corners consist of time surface values within a square $W \times W$ window centered at the detected corner.
In our implementation, we used $5 \times 5$ windows.

\subsection{Feature Matching and Management} \label{sec:matching}

To obtain a set of features that we would like to project in 3D to perform pose optimization, we implement both stereo and temporal feature matching, as depicted in Fig.~\ref{fig:ts}.
Features are stereo matched between the left and right time surface at the current timestamp $t$, $\tau_L(t)$ and $\tau_R(t)$, and temporally matched with the previous two time surfaces at the timestamp $t - \delta t$, $\tau_L(t-\delta t)$ and $\tau_R(t-\delta t)$.
Storing time surfaces from only one previous step makes our method very efficient in terms of memory.
Feature detection is done with raw event coordinates, but stereo rectified time surfaces are used for feature matching in order to ensure that the matching features from the left and right sensor are located along the epipolar lines.
The method has dynamic pose estimate frequency adapted to the rate of incoming events, meaning that $\delta t$ does not need to be predefined, but the number of events between estimates $N$ becomes a parameter.
The number of events used for pose optimization depends mainly on the resolution and the amount of texture in the scene.

Correspondences between events detected as corners are identified by leveraging the zero-normalized cross-correlation score between the feature descriptors -- vectors consisting of time surface values.
Although cross-correlation is more computationally expensive to calculate than some other common matching scores, like the sum of absolute differences, we find that using it drastically improves accuracy and thus justifies the added computational cost.
We dismiss the feature matches whose matching score is below a certain predefined threshold, set to $ZNCC_{min} = 20$ in our implementation, to retain only strong features.
To further increase quality of the feature set and filter out unreliable matches, we employ circular feature matching.
Feature matches will only be used for pose optimization if they have been matched through all four time surfaces by following a circular matching order: $\tau_L(t)$, $\tau_R(t)$, $\tau_R(t-\delta t)$, $\tau_R(t-\delta t)$, and then back to $\tau_L(t)$.
If the matching circle is closed with the first feature being matched to the same one at the end, we consider this to be a reliable feature.
Using only the selected subset of features not only speeds up computation during pose optimization, but also ensures more accurate results.

\begin{figure*}[h!]
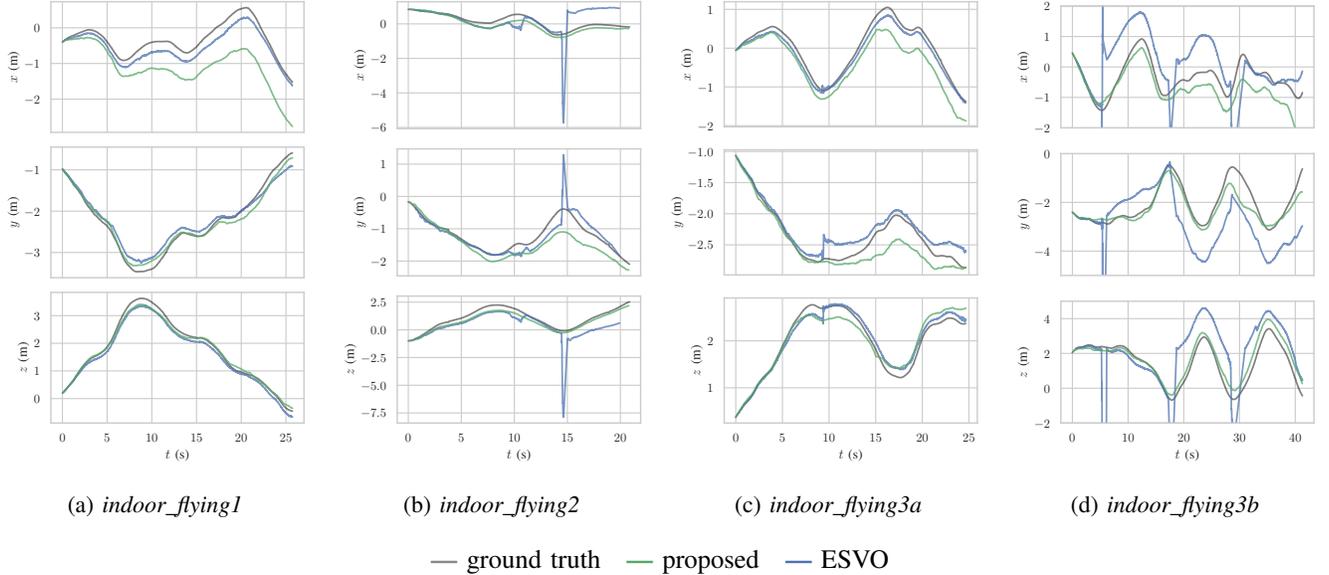

\centering
\begin{subfigure}{.24\textwidth}
  \resizebox{1.0\linewidth}{!}{\input{figures/plot_flying1_xyz_view.pgf}}
  \caption{\textit{indoor\_flying1}}
  \label{fig:mvsec1}
\end{subfigure}
\begin{subfigure}{.24\textwidth}
  \resizebox{1.0\linewidth}{!}{\input{figures/plot_flying2_xyz_view.pgf}}
  \caption{\textit{indoor\_flying2}}
  \label{fig:mvsec2}
\end{subfigure}
\begin{subfigure}{.24\textwidth}
  \resizebox{1.0\linewidth}{!}{\input{figures/plot_flying3_xyz_view.pgf}}
  \caption{\textit{indoor\_flying3a}}
  \label{fig:mvsec3}
\end{subfigure}
\begin{subfigure}{.24\textwidth}
  \resizebox{1.0\linewidth}{!}{\input{figures/plot_flying3_2_xyz_view.pgf}}
  \caption{\textit{indoor\_flying3b}}
  \label{fig:mvsec4}
\end{subfigure} \\ \vspace{0.4cm}
{\color{gray_graph}\textbf{\textemdash}} ground truth \hspace{0.1cm}
{\color{green_graph}\textbf{\textemdash}} proposed \hspace{0.1cm}
{\color{blue_graph}\textbf{\textemdash}} ESVO

\caption{Comparison of translation estimates of the proposed method and ESVO against ground truth provided by the motion capture system on the four subsequences of the MVSEC dataset. Ground truth is depicted in gray, our method in blue, and ESVO in green. Coordinates $x$, $y$, and $z$ are shown by rows. The spikes in ESVO results in (b) and (d) are caused by method  reinitialization. The figures show raw trajectories with matching origins.}
\label{fig:mvsec}
\end{figure*}

\subsection{Pose Optimization} \label{sec:pose}

Rotation and translation that correspond to the change in camera pose from the previous to the current view are jointly estimated by minimizing the sum of reprojection errors.
Feature points from the previous view are projected into 3D using the known camera calibration parameters.
Next, 3D points $ \mathbf{X} = (x, y, z)^\top $ are reprojected back to the 2D plane of the current image view at time $\mathbf{t}$ via the following equation:

\begin{equation}
\begin{bmatrix}
u \\
v \\
1
\end{bmatrix} =
\pi(\mathbf{X}; \mathbf{R}, \mathbf{t}) =
\begin{bmatrix}
f & 0 & c_u \\
0 & f & c_v \\
0 & 0 & 0
\end{bmatrix}
\begin{bmatrix}
\mathbf{R} | \mathbf{t}
\end{bmatrix}
\begin{bmatrix}
x \\
y \\
z \\
1
\end{bmatrix},
\end{equation}
where $(u, v)^\top$ are homogeneous feature coordinates, \mbox{$\pi \colon \R^3 \to \R^2$} is the projection function, $f$ is the focal length, $(c_u, c_v)^\top$ are the image principal point coordinates, and $\mathbf{R}$ and $\mathbf{t}$ are the rotation matrix and translation vector between the previous and current view, respectively.

Let $\pi^{(l)}$ be the projection of the 3D point onto the left image plane, and $\pi^{(r)}$ onto the right image plane.
With $\mathbf{x}_i^{(l)}$ and $\mathbf{x}_i^{(r)}$ being the feature coordinates in the current image view, the reprojection error equates to:

\begin{equation}
\sum_i \big\| \mathbf{x}_i^{(l)} - \pi^{(l)}(\mathbf{X}_i; \mathbf{R},  \mathbf{t})\big\|^2 + \big\| \mathbf{x}_i^{(r)} - \pi^{(r)}(\mathbf{X}_i; \mathbf{R}, \mathbf{t})\big\|^2.
\end{equation}
To find the transformations $\mathbf{R}$ and $\mathbf{t}$ that minimize the reprojection error, we apply the Gauss-Newton optimization method.
We also apply the RANSAC scheme, as suggested in \cite{geiger2011stereoscan}, by first estimating $\mathbf{R}$ and $\mathbf{t}$ for 50 times using 3 randomly chosen features.
All points are then tested against each of the 50 hypotheses, and the estimation with the most successful result is further refined by using all the inliers, i.e. the points whose distance of their reprojection to their matching 2D features is within a certain threshold.
Finally, we are able to integrate the $\mathbf{R}$ and $\mathbf{t}$ estimations to obtain full camera trajectory through time.

\section{Experimental Results} \label{sec:results}

\begin{figure*}[!t]
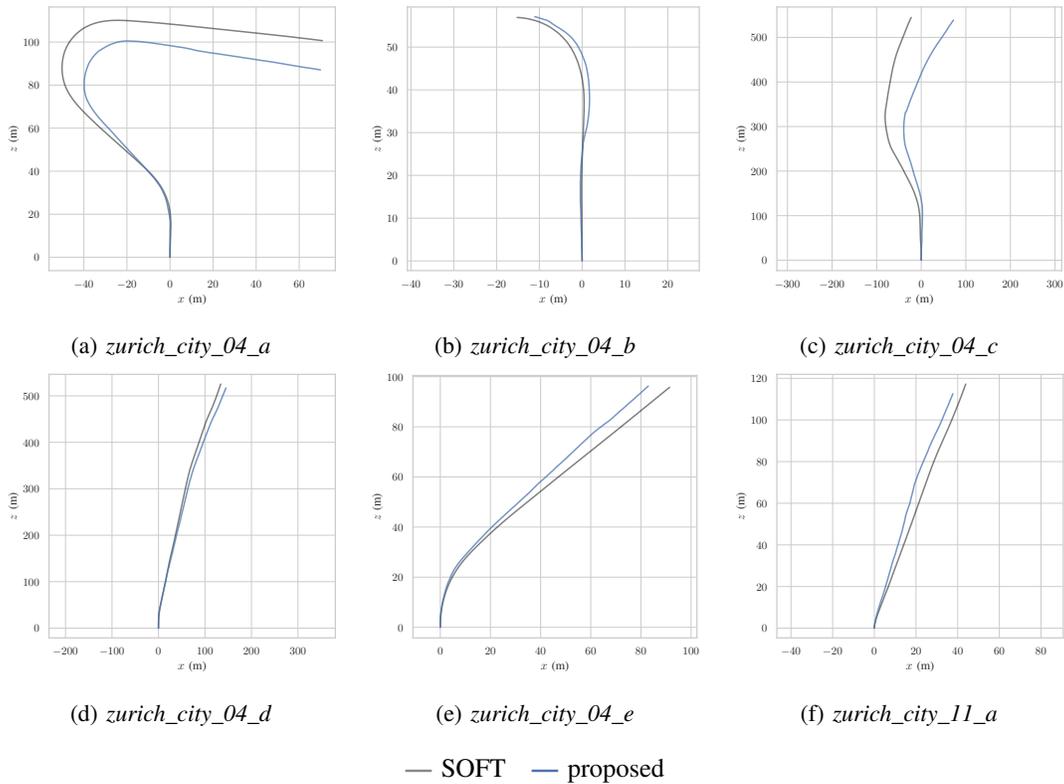

\centering
\begin{subfigure}{.26\textwidth}
  \resizebox{1.0\linewidth}{!}{\input{figures/zurich_a_trajectories.pgf}}
  \caption{\textit{zurich\_city\_04\_a}}
  \label{fig:dsec1}
\end{subfigure}
\begin{subfigure}{.26\textwidth}
  \resizebox{1.0\linewidth}{!}{\input{figures/zurich_b_trajectories.pgf}}
  \caption{\textit{zurich\_city\_04\_b}}
  \label{fig:dsec2}
\end{subfigure}
\begin{subfigure}{.26\textwidth}
  \resizebox{1.0\linewidth}{!}{\input{figures/zurich_c_trajectories.pgf}}
  \caption{\textit{zurich\_city\_04\_c}}
  \label{fig:dsec3}
\end{subfigure} \\
\begin{subfigure}{.26\textwidth}
  \resizebox{1.0\linewidth}{!}{\input{figures/zurich_d_trajectories.pgf}}
  \caption{\textit{zurich\_city\_04\_d}}
  \label{fig:dsec4}
\end{subfigure}
\begin{subfigure}{.26\textwidth}
  \resizebox{1.0\linewidth}{!}{\input{figures/zurich_e_trajectories.pgf}}
  \caption{\textit{zurich\_city\_04\_e}}
  \label{fig:dsec5}
\end{subfigure}
\begin{subfigure}{.26\textwidth}
  \resizebox{1.0\linewidth}{!}{\input{figures/zurich_11a_trajectories.pgf}}
  \caption{\textit{zurich\_city\_11\_a}}
  \label{fig:dsec6}
\end{subfigure} \\ \vspace{0.4cm}
{\color{gray_graph}\textbf{\textemdash}} SOFT \hspace{0.1cm}
{\color{blue_graph}\textbf{\textemdash}} proposed
\caption{2D trajectory estimates for the \textit{zurich\_city\_04} sequences from the DSEC dataset captured during outdoor driving. Results of frame-based visual odometry method SOFT used in lieu of ground truth are depicted in gray, and our method in blue. The figures show raw trajectories with aligned origins.}
\label{fig:dsec}
\end{figure*}

We provide experimental results for two different publicly available datasets: MVSEC \cite{zhu2018multivehicle} and DSEC \cite{gehrig2021dsec}.
Performance of our method is evaulated using the relative pose error defined as an average relative translation and rotation root-mean-square-error (RMSE) over a sliding window of different lengths using the RPG Trajectory Evaluation tool \cite{Zhang18iros}.
Figures \ref{fig:mvsec} and \ref{fig:dsec}, which present achieved results, are generated by the publicly available tool EVO \cite{grupp2017evo}.
We implemented our method in C++ in a ROS environment.
We witnessed real-time results on the MVSEC dataset using a laptop with Intel Core i7-7700HQ CPU at 2.80GHz on a single core.
However, due to very heavy event load in the DSEC dataset, we needed to slow down the \textit{rosbag} reproduction.
For the MVSEC dataset, we used $N=10000$ events from one sensor between pose estimates, and for the DSEC dataset, $N=150000$ events, as it has higher resolution and is very rich in texture, thus generating a lot of events.

\subsection{Experimental Data} \label{sec:data}

For experimental evaluation, we used data from the MVSEC dataset collected in an indoor environment with a flying drone carrying a stereo DAVIS346 event camera with a resolution of $346 \times 260$ pixels and a stereo baseline of 10 cm.
The drone mostly moves in translation with occasional hovering, thus we put an emphasis on the translation results.
Two partial sequences from the MVSEC dataset, \textit{indoor\_flying1} and \textit{indoor\_flying3a}, were used in the ESVO paper \cite{zhou2021event}, but herein we also add \textit{indoor\_flying2} and \textit{indoor\_flying3b} for more extensive evaluation.
These sequences offer ground truth poses provided by a motion capture system.
Even though there are also outdoor sequences captured by a driving car and motorcycle available for this dataset, we found that low spatial resolution of the cameras and small baseline of the stereo setup limits the performance on these sequences, especially in terms of the scale, thus we have not used these sequences in our evaluation.

Nevertheless, outdoor driving scenarios are available in the DSEC dataset.
It offers data from a stereo event camera mounted on a driving car with a higher resolution of $640 \times 480$ pixels and a baseline of 60~cm.
Although herein ground truth trajectories are not readily available, we make use of the available high-resolution ($1440 \times 1080$) standard frames and compare our method to a highly accurate frame-based visual odometry method SOFT \cite{cvisic2015stereo, cvisic2018soft} in lieu of ground truth.
For this dataset we chose a subset of sequences captured in the city during daylight. 
Even though it would be very interesting to see how event-based methods perform against frame-based in low light and high dynamic range environments, where one would expect event cameras to outperform the standard ones, unfortunately there are still no datasets published to allow this kind of experiments due to ground truth unavailability.

\subsection{Method Evaluation} \label{sec:eval}

We evaluate our method against ground truth on the MVSEC \textit{indoor\_flying} sequences and compare it to the only currently available stereo event-based odometry method, ESVO \cite{zhou2021event} that outputs pose estimates at a fixed predefined frequency (100 Hz in our experiments).
Given that, the ESVO system needs to reinitialize in case of too few events incoming (e.g., drone hovering or moving very slowly), but we only consider dynamic parts of the sequences to ensure fair comparison.
We use the parameters provided by the authors.
Translation results of our method compared to ESVO and ground truth on four partial sequences of the MVSEC dataset (upper rows) are shown in Fig.~\ref{fig:mvsec}, while relative translation and rotations errors are shown in Table \ref{table:results}.
We can see from the table that on sequences 1 and 3a ESVO is more accurate, especially in the translation.
However, our method does not suffer from sudden jumps in translation pose estimates in sequences 2 and 3b, as seen clearly in Figs.~\ref{fig:mvsec2} and \ref{fig:mvsec4}, respectively, thus also achieving lower error for both rotation and translation.
It is also important to note that in these two sequences total integrated length of the trajectory estimated by our method closely matches the ground truth one, while ESVO overestimates the length by around 100\%, which we believe is due to small oscillations around the mean estimated trajectory and high frequency.
We believe that rotation could be estimated better by our method if there were more events generated further away from the camera, but low spatial resolution and small baseline of the sensor are also a challenging factor.

Estimated trajectories for the DSEC dataset sequences \textit{zurich\_city\_04} and \textit{zurich\_city\_11\_a} are shown in Fig.~\ref{fig:dsec}.
We compare our results only to SOFT odometry, which is highly accurate on outdoor driving scenes, since we were unable to produce satisfying results for the ESVO method due to lack of suitable parameters.
From the relative pose errors shown in Table \ref{table:results}, we can see that the proposed method achieves good results in a natural environments rich in texture, even though there were not many events generated on the road near the car, which makes estimating translation more difficult.
In some of the segments, there were moving vehicles present in the significant part of the scene, e.g., in \textit{zurich\_city\_04\_b} depicted in Fig.~\ref{fig:dsec2}, which caused the odometry drift towards the right around the halfpoint of the trajectory.
A similar drift happened in \textit{zurich\_city\_04\_c} depicted in Fig.~\ref{fig:dsec3} at the beginning of the sequence.
We find that our method achieves fairly good results, even on very long driving sequences of up to more than 500~meters,  as it can be seen from the matching trajectories.

\begin{table}[!t]
\caption{Translation [\%] and rotation [\textdegree/m] evaluation results of the proposed visual odometry method compared to ESVO (where available) using relative pose error (RMSE).}
\begin{center}
{\renewcommand{\arraystretch}{1.5}
\begin{tabular}{  c  c c c c c c c}
 \hline
 & \multicolumn{2}{c}{Proposed} & \multicolumn{2}{c}{ESVO} & \multicolumn{3}{c}{Trajectory length [m]} \\
 \cline{2-8}
 \multicolumn{1}{c}{Data} & $\mathbf{t}$ & $\mathbf{R}$ & $\mathbf{t}$ & $\mathbf{R}$ & GT & Prop. & ESVO \\ \hline
  fly\_1    & 14.24 & 2.37 & 7.38 & 1.23 & 11.67 & 11.36 & 23.38  \\
  fly\_2    &  11.69 & 4.36 & 43.10 & 8.53 & 10.63 & 9.78 & 45.10 \\
  fly\_3a   &  14.03 & 2.44 & 8.25 & 1.47 & 10.57 & 10.19 & 23.79 \\
  fly\_3b   &  8.63 & 1.93 & 34.81 & 3.41 & 28.61 & 26.88 & 197.5 \\
  \hline
  04\_a &  7.59 & 0.10 & n/a & n/a & 241.0 & 218.5 & n/a \\
  04\_b & 8.03 & 0.20 & n/a & n/a  & 65.4 & 62.8 & n/a \\
  04\_c & 6.95 & 0.04  & n/a & n/a & 570.0 & 574.1 & n/a \\
  04\_d & 2.90 & 0.03  & n/a & n/a & 544.1 & 542.9 & n/a \\
  04\_e & 9.18 & 0.10  & n/a & n/a & 136.3 & 131.3 & n/a \\
  11\_a & 4.33 & 0.12  & n/a & n/a & 125.3 & 119.5 & n/a \\

 \hline
\end{tabular}}
\end{center}
\label{table:results}
\end{table}

\section{Conclusion} \label{sec:conclusion}

In this paper, we have presented a novel stereo visual odometry method for event cameras.
The method is based on feature detection with stereo and temporal matching, as well as careful feature selection to enable reliable pose estimation by reprojection error minimization.
The performance of the proposed method is quantitatively evaluated on two publicly available datasets, MVSEC captured by a flying drone, and on outdoor driving sequences from the DSEC dataset.
We compare our method to another stereo event-based visual odometry, namely ESVO, on the MVSEC dataset.
In contrast to ESVO, our method does not have a fixed tracking frequency and does not require initialization, and thus does not yield unreliable estimates in case of too few incoming events.
On the DSEC dataset, due to lack of ground truth, we evaluated our method with respect to an accurate frame-based odometry SOFT that is one of the highest ranking visual odometry algorithms on the KITTI dataset.
Our method achieved fairly good results on driving sequences of up to more than 500 m.
Future work will include fusing information from events and standard frames in a stereo setup, as well as the IMU, in order to build a more robust method that will exploit the complementary advantages of those sensors.

\balance

\bibliography{library}
\bibliographystyle{IEEEtran}

\end{document}